\documentclass[orivec]{llncs}
\usepackage[T1]{fontenc}
\usepackage[table]{xcolor}

\usepackage{multirow}
\usepackage{booktabs}
\usepackage{textcomp}
\usepackage{graphicx}
\usepackage{acronym}
\usepackage[most]{tcolorbox}
\usepackage{listings}
\usepackage{float}
\usepackage[table]{xcolor}

\newacro{LLM}{Large Language Model}
\newacro{MIDAS}{Multi-LLM Iterative Data-Adaptive Summarization}
\newacro{NLP}{Natural Language Processing}
\newacro{ROUGE}{Recall-Oriented Understudy for Gisting Evaluation}
\newacro{OEM}{Original Equipment Manufacturer}
\newacro{CriSPO}{Multi-Aspect Critique-Suggestion-guided Automatic
Prompt Optimization for Text Generation}
\newacro{CoT}{Chain-of-Thought}
\newacro{ICL}{In-Context Learning}

\usepackage{tcolorbox}
\tcbuselibrary{breakable, skins, listings}
\usepackage{amsmath}
\definecolor{policyblue}{RGB}{220,235,255}
\definecolor{policybluedark}{RGB}{30,90,180}
\definecolor{criticgreen}{RGB}{220,245,220}
\definecolor{criticgreendark}{RGB}{30,140,60}
\definecolor{scorerorange}{RGB}{255,240,220}
\definecolor{scorerorangedark}{RGB}{200,100,20}
\definecolor{rulenum}{RGB}{100,100,100}
\definecolor{midasrow}{RGB}{245,245,245}

\tcbset{
    midas base/.style={
    breakable, enhanced,
    fonttitle=\bfseries\small,
    coltitle=white,
    attach boxed title to top left={yshift=-2mm, xshift=4mm},
    boxed title style={rounded corners, size=small},
    arc=4pt, boxrule=0.8pt,
    left=6pt, right=6pt, top=10pt, bottom=6pt,
  },
  policy/.style={
    midas base,
    colback=policyblue!40,
    colframe=policybluedark,
    boxed title style={colback=policybluedark},
    title={Policy Block~---~\texttt{Sum\_Type2} (Enterprise IT help desk Tickets)},
  },
  critic/.style={
    midas base,
    colback=criticgreen!40,
    colframe=criticgreendark,
    boxed title style={colback=criticgreendark},
    title={Unified Critic Template~---~Input Structure},
  },
  scorer/.style={
    midas base,
    colback=scorerorange!50,
    colframe=scorerorangedark,
    boxed title style={colback=scorerorangedark},
    title={Scorer LLM~---~Evaluation Dimensions},
  },
}
\newcommand{\prule}[2]{%
  \noindent\textcolor{rulenum}{\textbf{#1.}}\hspace{4pt}#2\par\smallskip}
\begin{document}

\title{MIDAS: Multi-LLM Iterative Data-Adaptive Summarization}
\titlerunning{MIDAS}
%
%
%
%
\author{Karen Lee \and
Dhanashree Balaram \and
Seojun Shon \and Umair Rasheed\thanks{Project Manager}}

\institute{Volkswagen Group Innovation, California}

\maketitle              

\section{Abstract}
Text summarization is deceptively difficult. While condensing information seems straightforward, real-world enterprise summarization of support tickets, legal documents, incident reports, and more, demands strict adherence to domain-specific guidelines, output formats, and organizational conventions. Crafting prompts that reliably satisfy these constraints is labor-intensive, requiring significant human expertise and continuous maintenance as requirements evolve.
Existing automated prompt optimization methods reduce this burden through \ac{LLM} critique-driven refinement, yet remain limited by static prompts that cannot adapt to the diversity of summary applications. We propose \ac{MIDAS}, a multi-LLM framework that extends this paradigm with data-driven pattern learning and use-case-specific personalization, enabling automatic adaptation to different summarization requirements without manual prompt engineering.
Applied to enterprise customer ticket summarization across five output formats, \ac{MIDAS} achieves the strongest overall performance against state-of-the-art critique-driven optimization frameworks such as CriSPO and ZERA, improving ROUGE-1 by up to 11.0\%, ROUGE-2 by up to 18.2\%, and ROUGE-L by up to 8.0\%, while consistently improving BERTScore F\textsubscript{1} across all formats and output types. We additionally demonstrate cross-model and cross-domain generalization through multi-LLM configurations and finance-domain summarization benchmarks.

\section{Introduction}

Automated text summarization has matured significantly with the advent of \acp{LLM}, yet
deploying these systems in enterprise settings reveals a fundamental tension: different
organizations, teams, and use cases demand fundamentally different summaries. A legal
department summarizing contract clauses, a financial analyst condensing earnings reports,
and a logistics coordinator reviewing shipment incidents each expect distinct formats,
vocabularies, and levels of detail. This diversity makes one-size-fits-all prompt design
impractical at enterprise scale.

A particularly relevant domain is customer support ticketing. Across industries, from automotive \acp{OEM} and dealerships to software vendors and consumer electronics manufacturers, support tickets are the primary record of customer issues and resolutions.
At scale, manually processing these tickets is estimated to absorb 20--40\% of support agent
capacity~\cite{forrester_tei_zendesk_2023}. Summarization requirements vary considerably
across roles: a field service engineer needs technical root cause and diagnostics, while a
warranty administrator prioritizes customer impact and resolution status. These differences
create strict, role-specific expectations for format and content that generic prompting
strategies cannot reliably satisfy.

Existing prompt optimization frameworks such as Critique-Suggestion-guided Automatic Prompt Optimization for Text Generation(CriSPO)~\cite{He2024CriSPOMC} partially
address this through critique-driven iterative refinement, but apply critique dimensions
that are agnostic to the target dataset and do not leverage structural patterns present in
reference summaries. As a result, prompts must still be manually tuned per domain, limiting
scalability.

To address these limitations, we propose \textbf{\ac{MIDAS}}, a multi-\ac{LLM}
framework that grounds prompt optimization in the properties of the target domain. \ac{MIDAS}
analyzes reference summaries to extract domain-specific patterns and conditions critique generation on a
rich multi-source context, enabling automatic adaptation to diverse summarization requirements without manual prompt engineering. We evaluate \ac{MIDAS} on a large-scale enterprise IT help desk dataset~\cite{bueck_multilingual_2026} across five output format configurations, demonstrating consistent improvements over zero-shot, \ac{ICL}, and \ac{CriSPO} baselines.
\noindent Our main contributions are:
\begin{itemize}
    \item \textbf{Data-Aware Policy Block Generation:} A dedicated Data Pattern \ac{LLM}
    extracts domain-specific formatting constraints from reference summaries and encodes
    them as policy blocks that ground iterative prompt refinement.
    \item \textbf{Unified \ac{CoT} Critic:} A single \ac{CoT} \ac{LLM} jointly generates
    structured critique and a refined prompt in one pass, eliminating the separate suggestion
    step in \ac{CriSPO} and reducing inference overhead.
    \item \textbf{Comprehensive Evaluation:} Five output format configurations spanning
    structured metadata, subject lines, multilingual summaries, and keyword tags, with
    \ac{ROUGE}-1/2/L gains of up to \textbf{11.0\%/18.2\%/8.0\%} and consistent BERTScore
    F$_1$ improvements over \ac{CriSPO} across all formats.
    \item \textbf{Cross-Model and Cross-Domain Generalization:} MIDAS demonstrates consistent performance across heterogeneous LLM backbones and distinct enterprise domains, achieving the strongest overall performance against CriS-PO and ZERA across both IT help desk and finance summarization benchmarks.
\end{itemize}

\noindent The remainder of the paper is organized as follows. Section~\ref{sec:related}
reviews related work. Section~\ref{sec:method} describes the \ac{MIDAS} framework.
Section~\ref{sec:exp} presents the data and experimental setup, and Section~\ref{sec:results}
reports results and discussion. Section~\ref{sec:conclusion} concludes.
\section{Related Work}
\label{sec:related}

Automatic prompt engineering has emerged as a critical research direction for improving large language model (LLM) \cite{zhao2023survey} performance without the computational overhead of fine-tuning. We review existing approaches in two key dimensions: (1) automatic prompt optimization methods and (2) LLM personalization techniques. Our analysis reveals a systematic gap: while prompt optimization methods remain \emph{static in their adaptation criteria}, and personalization methods focus exclusively on \emph{individual user preferences}, neither addresses the challenge of learning \emph{organizational format requirements} from reference data.

\subsection{Automatic Prompt Optimization}

Early work on automatic prompt engineering established LLMs as effective optimizers of their own prompts. APE(Automatic Prompt Engineer) \cite{zhou2023ape}, OPRO (Optimization by PROmpting) \cite{yang2024opro}, and EvoPrompt \cite{guo2024evoprompt} demonstrate that LLMs can iteratively optimize prompts through candidate generation, trajectory-aware refinement, and evolutionary search strategies. However, these methods primarily rely on scalar optimization feedback rather
than interpretable structural guidance. This becomes limiting in enterprise summarization settings where "quality" is not only semantic correctness but also adherence to organization-specific templates, headings, and prefixes.

\subsection{Critique and Feedback Guided Prompt Refinement}

To make optimization more interpretable and actionable, recent work incorporates natural language critique as an intermediate signal. CriSPO \cite{He2024CriSPOMC} proposes Critique-Suggestion-guided Prompt Optimization, designed specifically for text generation tasks where metrics like \ac{ROUGE} \cite{lin2004rouge} provide limited guidance. CriSPO introduces a critique-guided prompt refinement for text generation by producing actionable natural-language feedback across multiple evaluation aspects. However, its critique policy remains \textbf{static} and does not adapt to domain-specific structural conventions from reference summaries.

ZERA \cite{yi2025zera} also follows the critique-driven lineage but grounds refinement in eight predefined, task-agnostic evaluation principles—completeness, conciseness, correctness, expression style, faithfulness, meaning accuracy, reasoning quality, and structural alignment—whose relative importance weights are inferred per task. It separates evaluation (principle-based critique generation) from refinement (meta-cognitive prompt refinement) and jointly optimizes the system prompt, user prompt, and task description from an underspecified ("zero-init") initialization using few samples and short iteration cycles. Although ZERA improves generality through principle-based optimization, its
evaluation rubric remains task-agnostic and does not explicitly model dataset-
specific formatting constraints such as mandatory prefixes, delimiters, or
language normalization rules. MIDAS complements this approach through a
dedicated Data Pattern LLM that derives such constraints directly from
reference data via policy block induction.

Recent approaches such as ProRefine \cite{weerasooriya2025prorefine}, PDO (Prompt-Dueling Optimization) \cite{chen2025pdo}, and PMPO (Probabilistic Metric Prompt Optimization) \cite{wang2025pmpo} explore alternative prompt refinement strategies including inference-time feedback loops, pairwise prompt comparison, and probabilistic metric-based optimization. While these methods improve refinement efficiency and search stability, they do not directly address the core requirement in enterprise summarization: \textbf{discovering and enforcing organization-specific output schemas} induced from reference data.

\subsection{LLM Personalization}

Personalization research aims to adapt LLM behavior to individuals by modeling user preferences, writing style, or interests. A common approach is \textbf{retrieval-augmented personalization}, where past user interactions are retrieved and injected into prompts at runtime. LaMP~\cite{salemi2024lamp} formalizes this setting through a benchmark suite spanning classification and generation tasks with user profiles and histories, while summary-augmented retrieval approaches~\cite{richardson2023integratingsummarizationretrievalenhanced} combine offline user summaries with selective retrieval to reduce retrieval overhead while preserving personalization quality.

Beyond discrete prompting, \textbf{soft prompt} methods encode user history into continuous embeddings that function as learned prompt vectors. Representative approaches include PERSOMA~\cite{hebert2024persoma}, PeaPOD~\cite{ramos2024peapod}, CoLLM~\cite{zhang2025collm}, and PersonalLLM~\cite{zollo2025personalllm}, which model personalization through soft prompts, collaborative embeddings, or reward-model ensembles.

Existing personalization approaches primarily target \textbf{user-level adaptation} such as writing style, recommendation behavior, or preference alignment, often relying on retrieval pipelines, embedding methods, or parameter-efficient tuning. In contrast, enterprise summarization requires \textbf{organizational-level personalization}: adherence to organization-specific templates, terminology, and formatting conventions. MIDAS addresses this setting through a purely prompt-based framework that learns structural constraints directly from reference data via data-driven pattern learning and policy block induction.

\section{Methodology}
\label{sec:method}

\begin{figure}[htbp]
\centering
\includegraphics[width=\textwidth]{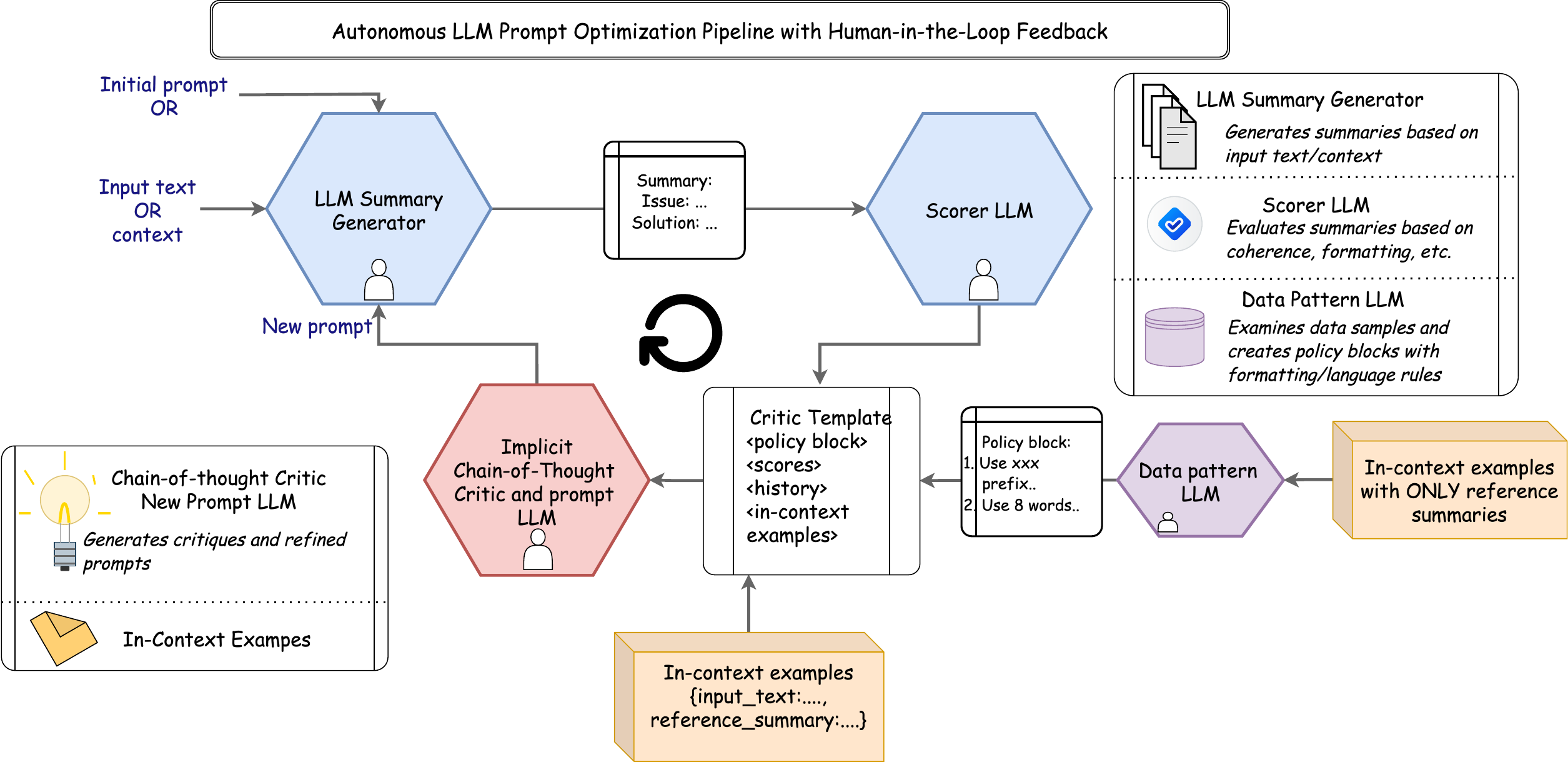}
\caption{Overview of MIDAS, an iterative multi-agent framework for automated prompt refinement. An LLM Summary Generator produces summaries that are evaluated by a Scorer LLM. A Data Pattern LLM extracts dataset-specific structural rules from a stratified holdout subset. These signals, together with optimization history, are provided to an Implicit Chain-of-Thought Critic and Prompt LLM, which generates critiques and refined prompts. The loop repeats for $N$ iterations, and the best-performing prompt is selected for final inference.}
\label{fig:method}
\end{figure}


\subsection{Data Pattern Learning}

 MIDAS introduces a Data Pattern Learning component implemented via in-context learning. A dedicated subset of the training data is reserved exclusively for structural pattern extraction (ensuring no leakage into train or test set for optimization).
Rather than random sampling, we adopt a distribution-aware sampling strategy to preserve key structural proportions in the dataset. Reference summaries (ground-truth summaries paired with each input ticket) are analyzed to identify formatting regularities. For example, if 30\% of reference summaries begin with a specific prefix (e.g., ``Severity:''), the holdout subset maintains this proportion, ensuring the extracted patterns faithfully reflect the full dataset distribution rather than a biased sample.
The sampled reference summaries are then provided to a dedicated \textbf{Data Pattern LLM}, which identifies structural regularities such as formatting rules, section ordering, and stylistic constraints, and encodes them as a set of explicit natural-language rules as a \textit{policy block}. The exact rule-induction template used for policy block generation is provided in Appendix~\ref{app:data_pattern_prompt}. For example, a customer ticket dataset might create rules such as: \textit{``Summaries must begin with a Severity field'',} or \textit{``Resolution steps should be listed as a numbered sequence''}. These rules collectively form a \textbf{policy block} that is injected into the critic template, grounding downstream critique and prompt refinement in the observed conventions of the target domain.

\begin{tcolorbox}[policy]
\footnotesize
Policy block for \texttt{Summary Format 2} (enterprise IT help desk dataset), illustrating the
two-tier rule structure used in MIDAS.

\medskip
\noindent\textcolor{policybluedark}{\textbf{Static Rules}}

\medskip
\prule{1}{\textbf{Output format.}
  Every output must use the exact four-field one-line format with commas and spaces as shown:\\
  \texttt{Type: X, Queue: Y, Priority: Z, Language: L}}

\prule{2}{\textbf{Valid ticket types \& priority levels.}
  \texttt{Type} must be exactly one of:
  \texttt{Incident}, \texttt{Request}, \texttt{Change}, \texttt{Problem}. \texttt{Priority} must be exactly one of: \texttt{high}, \texttt{medium}, \texttt{low}.}

\prule{3}{\textbf{No additional text.}
  Output only the four fields with no explanation, preamble, or trailing content.}

\medskip
\noindent\textcolor{criticgreendark}{\textbf{Data-Learned Rules examples}}

\medskip
\prule{4}{\textbf{Incident classification.}
  When the ticket describes a system issue, malfunction, outage, crash, downtime, or
  unauthorized access $\rightarrow$ \texttt{Type: Incident}.\\
  \textit{Example}: \texttt{``server overload and subsequent downtime''} $\rightarrow$ \texttt{Type: Incident}}

\prule{5}{\textbf{Queue routing --- Billing.}
  When billing, invoices, charges, or payment systems are the main topic
  $\rightarrow$ \texttt{Queue: Billing and Payments}.}

\prule{6}{\textbf{Priority inference.}
  When the ticket signals urgency or high impact via keywords such as
  \textit{``critical''}, \textit{``breach''}, \textit{``unauthorized access''}, or
  \textit{``significantly affecting productivity''} $\rightarrow$ \texttt{Priority: high}.\\
  \textit{Example}: \texttt{``A critical outage has been reported\ldots''} $\rightarrow$ \texttt{Priority: high}}

\prule{7}{\textbf{Language detection.}
  Detect the input language and set \texttt{Language} accordingly.
  }
  
\end{tcolorbox}

\subsection{Multi-LLM Summary Generation and Implicit \ac{CoT} Critic and Prompt Optimization} \label{sec:midas_multi_llm}

MIDAS implements an iterative prompt refinement loop using three coordinated LLM agents: (1) an \textbf{LLM Summary Generator}, (2) a \textbf{Scorer LLM}, and (3) an \textbf{Implicit Chain-of-Thought Critic and Prompt LLM} (Figure~\ref{fig:method}). In our implementation, all three agents share the same underlying foundation model, instantiated with different prompts tailored to their respective roles. The framework is model-agnostic and can, in principle, be instantiated with alternative backbone models. For stability, we use deterministic decoding for scoring-time generation to reduce evaluation variance.

\paragraph{Summary Generation.}
Given an input ticket $x$ and a task prompt $p$ (initialized with an initial prompt and updated across iterations; see Appendix~\ref{app:init_prompt}), the LLM Summary Generator produces a summary $\hat{y}=\mathrm{Gen}(x,p)$. Concretely, we generate one summary per input in the selected evaluation subset, where the input to the generator is the raw ticket text formatted with the current prompt. Across iterations, $p$ is updated by the prompt refinement module (described below).

\paragraph{LLM-Based Evaluation (Scorer LLM).}
To score generated summaries, MIDAS uses an LLM-based evaluator rather than relying solely on automatic metrics such as ROUGE~\cite{lin2004rouge} or BERTScore~\cite{Zhang2019BERTScoreET}. Prior work~\cite{nguyen2024comparative} suggests that LLM-based evaluators better align with human judgment on summarization tasks. Our Scorer LLM evaluates a generated summary $\hat{y}$ against a reference summary $y$ given the original ticket $x$, using a structured rubric that returns multiple dimension scores (e.g., \emph{Core Meaning}, \emph{Unsupported Additions}, \emph{Format \& Style Fidelity}) and an explanation. The full evaluation prompt and scoring rubric used by the Scorer LLM are reproduced verbatim in Appendix~\ref{app:scorer_prompt}. We then compute a scalar optimization score as a weighted combination of these dimensions, with weights configurable by the user:
\[
s(\hat{y},y,x)=\sum_{d \in \mathcal{D}} w_d \, s_d(\hat{y},y,x), 
\quad \text{where } \sum_d w_d = 1.
\]
In our experiments, we set $(w_{\text{core}}, w_{\text{unsupported}}, w_{\text{format}})=(0.4,0.3,0.3)$, based on our task-specific prioritization of semantic fidelity and formatting consistency; however, these guidelines are user-configurable and can be adjusted to reflect different application requirements. For stability, scoring uses deterministic generation where applicable.

\paragraph{Multi-suggestion Prompt Proposal and Fast Selection.}
At each iteration, users may request $M$ candidate prompt suggestions. For each candidate prompt, MIDAS performs a fast evaluation on a \emph{representative subset} of the training set (default: 25 examples out of 40) to estimate the candidate's average LLM-based score. The subset is selected using the same distribution-aware sampling strategy described in Section~4.1.

\paragraph{Implicit Chain-of-Thought Critic and Prompt LLM.}
The prompt refinement stage jointly produces (i) a structured critique of the current prompt behavior and (ii) a revised prompt for the next iteration. We implement this using a single LLM call with two tagged outputs: \texttt{<Critique>} and \texttt{<Suggestion>}, where the suggestion contains the full revised task prompt. The critic template incorporates the current prompt and score, generated examples (raw text, generated summary, reference summary), optimization history consisting of the top-$K$ prior prompts with associated scores and critiques, and a policy block containing dataset-specific formatting constraints induced by the Data Pattern LLM (Section~4.1). This policy block grounds critique and refinement in observed domain conventions. The complete critic template is provided in Appendix~\ref{app:critic_prompt_template}, and representative prompt templates are included in Appendix~\ref{app:prompts}.

\paragraph{Efficiency Relative to CriSPO.}
A key architectural difference from \ac{CriSPO}~\cite{He2024CriSPOMC} is that
\ac{MIDAS} consolidates critique generation and prompt optimization into a
single module (Implicit Chain-of-Thought Critic and Prompt LLM). \ac{CriSPO}
separates these into two LLM stages: one LLM produces critiques and suggestions,
while another consumes those suggestions to generate the next prompt. In
\ac{MIDAS}, critique signals are directly incorporated into prompt refinement
within a single generation step, eliminating an intermediate LLM call and
streamlining the iterative optimization process.

\section{Experiments}
\label{sec:exp}

\subsection{Dataset}
\label{sec:data}

We evaluate \ac{MIDAS} on a publicly available multilingual enterprise IT help desk ticket dataset~\cite{bueck_multilingual_2026}, containing approximately 50,000 support tickets spanning multiple organizational workflows, languages, and service queues. Each record contains customer ticket content (\textit{Subject} and \textit{Body}), the corresponding help desk response (\textit{Answer}), and associated structured metadata such as ticket type, routing queue, priority, language, business category, and categorical tags.

After filtering for entries with all required fields present across all five output configurations, we obtain a working corpus of \textbf{24,635 tickets}. All experiments use \textbf{GPT-5.2} as the backbone model for summary generation, critique, scoring, and data pattern learning.

\paragraph{Data Representation.}
Each ticket is converted into a structured input concatenating the body and agent answer and paired with five parallel reference summaries. Table~\ref{tab:data_example} shows a
representative example.

\begin{table}[ht]
\centering
\caption{Example of a single ticket converted into all five output format configurations.
\texttt{Format Type3} and \texttt{Format Type4} reference summaries are generated by GPT-5.2
conditioned on the ticket's tag fields (grounded generation), and subsequently verified
by human annotators to ensure factual consistency with the source ticket.}
\label{tab:data_example}
\small
\begin{tabular}{@{} l p{0.72\linewidth} @{}}
\toprule
\textbf{Field} & \textbf{Content} \\
\midrule
\textbf{Raw Text}       & \texttt{Body: Dear Customer Support Team, I am writing to report a} \\
                        & \texttt{significant problem with our account\ldots Answer: \ldots} \\
\midrule
\textbf{Format Type1}     & \texttt{Type: Incident, Queue: Technical Support, Priority: high, Language: en} \\
\textbf{Format Type2}     & \texttt{Subject: Account Disruption} \\
\textbf{Format Type3}     & \texttt{There is an account outage causing a disruption.} \\
\textbf{Format Type4}     & \texttt{There is an account outage causing a disruption.} \\
\textbf{Format Type5}     & \texttt{Tag\_1: Account, Tag\_2: Disruption, Tag\_3: Outage} \\
\bottomrule
\end{tabular}
\end{table}

\noindent Note that \texttt{Format Type3} and \texttt{Format Type4} are identical for English-language
tickets; the distinction becomes apparent for non-English inputs, where \texttt{Format Type3}
mirrors the ticket's original language while \texttt{Format Type4} normalizes to English.

\paragraph{Output Format Configurations.}
Different enterprise roles require different summary structures from the same underlying ticket. To reflect this, we construct five output format configurations (Format Type1--Format Type5), where each ticket is paired with one of five reference summary formats:

\begin{itemize}
    \item \textbf{\texttt{Format Type1} --- Structured Metadata.}
    \texttt{Type:\{type\}, Queue:\{queue\}, Priority:\{priority\}, Language:\{language\}}.

    \item \textbf{\texttt{Format Type2} --- Subject Line.}
    \texttt{Subject:\{subject\}}.
    A concise one-line summary. Entries with missing subject fields (3,838) are excluded from this configuration.

    \item \textbf{\texttt{Format Type3} --- Natural Language Summary (Original Language).}
    A free-text summary generated by GPT-5.2 from ticket tag metadata, then verified and corrected by human annotators. Summaries are written in the original language of the ticket to evaluate multilingual generation behavior.

    \item \textbf{\texttt{Format Type4} --- Natural Language Summary (English Only).}
    The same summary format as \texttt{Format Type3}, but normalized to English, testing whether \ac{MIDAS} can learn and enforce language normalization constraints from reference summaries.

    \item \textbf{\texttt{Format Type5} --- Structured Tag Output.}
    \texttt{tag\_1: \{tag\_1\}, tag\_2: \{tag\_2\}, tag\_3: \{tag\_3\}}.
    Only the first three tag fields are used, as later tag columns are sparsely populated in the filtered corpus.
\end{itemize}

\subsection{Experimental Setup}
\label{sec:setup}

\paragraph{Data Splits.}
For each output configuration, prompt optimization is performed using a fixed pool of \textbf{40 training examples} and \textbf{10 development examples}. We analyze performance on \textbf{30 test samples} drawn randomly from the remaining corpus, with the same random seed applied across all configurations to ensure comparability. A separate stratified holdout subset of \textbf{200 samples} is reserved exclusively for data pattern learning and is excluded from both optimization and evaluation to prevent data leakage. Beyond these splits, the remaining \textbf{24,355 tickets} serve as a large-scale unseen holdout set, used to validate the best-performing prompts from each experimental condition and confirm that gains generalize beyond the small optimization pool. \\

We compare \ac{MIDAS} against Zero-Shot, \ac{ICL} ($k{=}3$), \ac{CriSPO}, \ac{CriSPO}
(100 iter), and ZERA, a state-of-the-art framework for critique-driven prompt optimization. We report \ac{ROUGE}-1/2/L F$_1$~\cite{Lin2004ROUGEAP} and BERTScore
F$_1$~\cite{Zhang2019BERTScoreET} independently for each output type. \ac{ROUGE} is
particularly informative for structured outputs (\texttt{Format~1}, \texttt{Format~5})
where exact field matching is expected, while BERTScore better captures semantic similarity
for free-text configurations (Format 3, Format 4). All runs use
$N{=}30$ optimization iterations (selected via ablation over $N \in \{10, 20, 30, 50\}$;
Figure~\ref{fig:opt_plot}) with identical generation parameters and 2 random seeds.

Our experiments address six questions: whether \ac{MIDAS} achieves the strongest overall performance across output formats (Table~\ref{tab:main_results_gpt52}); whether its gains
are architectural rather than a product of more iterations (\ac{CriSPO} 100 iter controls
for compute budget); what the optimal iteration count is (Figure~\ref{fig:opt_plot}); whether \ac{MIDAS} generalizes across structurally diverse output types spanning metadata,
subject lines, multilingual summaries, and keyword tags; whether the framework remains effective under heterogeneous multi-LLM configurations (Table~\ref{tab:main_results_gpt52}); and whether the proposed policy-learning mechanism generalizes across distinct enterprise domains (Table~\ref{tab:finance_results}).

\subsection{Cross-Domain Evaluation} To evaluate whether \ac{MIDAS} generalizes beyond enterprise IT help desk summarization, we additionally evaluate on the ECTSum finance-domain benchmark, which consists of financial earnings call transcripts paired with concise analyst-style summary annotations. Compared to the IT help desk dataset, ECTSum contains substantially different terminology, discourse structure, and summarization objectives, focusing on financial performance indicators, operational reporting, and market-related events rather than incident resolution workflows. For cross-domain evaluation, we compare \ac{MIDAS} against \ac{CriSPO} and ZERA under the same GPT-5.2 backbone and identical 30-iteration optimization setting used in the primary experiments. This experiment evaluates whether data-driven policy learning generalizes under domain shift without requiring manually engineered formatting rules.
\section{Results and Discussion}
\label{sec:results}

\begin{table}[!htbp] \centering \caption{Evaluation results on the \textbf{enterprise IT help desk dataset} for Zero-Shot (ZS), In-Context Learning (ICL, $k{=}3$), CRISPO, CRISPO (100 iter), ZERA, and two MIDAS instantiations across all five summary output types. MIDAS uses GPT-5.2 for all agents, while MIDAS (Multi-LLM) uses heterogeneous backbone assignments. R-1/2/L denote ROUGE-1/2/L F$_1$; BS-F1 denotes BERTScore F$_1$. Best result per metric per type is in \textbf{bold}.} \label{tab:main_results_gpt52} \definecolor{midasrow}{RGB}{245,245,245} \begin{tabular}{@{} l l cccc @{}} \toprule \textbf{Summary} & \textbf{Method} & \textbf{R-1} & \textbf{R-2} & \textbf{R-L} & \textbf{BS-F1} \\ \midrule \multirow{7}{*}{\texttt{Format 1}} & ZS & 0.6655 & 0.3928 & 0.6654 & 0.9530 \\ & ICL & 0.7330 & 0.4876 & 0.7330 & 0.9583 \\ & CRISPO & 0.7352 & 0.4768 & 0.7352 & 0.9673 \\ & CRISPO (100 iter) & 0.7154 & 0.4514 & 0.7154 & 0.9168 \\ & ZERA & \textbf{0.7549} & \textbf{0.5353} & \textbf{0.7549} & 0.9676 \\ \rowcolor{midasrow} & MIDAS & 0.7521 & 0.5303 & 0.7521 & \textbf{0.9688} \\ \rowcolor{midasrow} & MIDAS (Multi-LLM) & 0.7312 & 0.5035 & 0.7312 & 0.9653 \\ \cmidrule(lr){1-6} \multirow{7}{*}{\texttt{Format 2}} & ZS & 0.2789 & 0.1240 & 0.2499 & 0.8101 \\ & ICL & 0.4364 & 0.1670 & 0.4088 & 0.8632 \\ & CRISPO & 0.4565 & 0.1431 & 0.4146 & 0.8715 \\ & CRISPO (100 iter) & 0.4552 & 0.1640 & 0.4206 & 0.8743 \\ & ZERA & 0.4350 & 0.1458 & 0.4099 & 0.8646 \\ \rowcolor{midasrow} & MIDAS & \textbf{0.4744} & 0.1699 & \textbf{0.4431} & 0.8800 \\ \rowcolor{midasrow} & MIDAS (Multi-LLM) & 0.4669 & \textbf{0.1833} & 0.4408 & \textbf{0.8801} \\ \cmidrule(lr){1-6} \multirow{7}{*}{\texttt{Format 3}} & ZS & 0.3654 & 0.1776 & 0.2932 & 0.8351 \\ & ICL & 0.4418 & 0.2268 & 0.3665 & 0.8554 \\ & CRISPO & 0.3749 & 0.1862 & 0.3080 & 0.8361 \\ & CRISPO (100 iter) & 0.4211 & 0.2065 & 0.3410 & 0.8528 \\ & ZERA & 0.4523 & 0.2451 & 0.3807 & 0.8574 \\ \rowcolor{midasrow} & MIDAS & \textbf{0.4862} & \textbf{0.2795} & \textbf{0.4253} & 0.8693 \\ \rowcolor{midasrow} & MIDAS (Multi-LLM) & 0.4822 & 0.2668 & 0.4126 & \textbf{0.8707} \\ \cmidrule(lr){1-6} \multirow{7}{*}{\texttt{Format 4}} & ZS & 0.4364 & 0.2094 & 0.3425 & 0.8606 \\ & ICL & 0.5111 & 0.2509 & 0.4131 & 0.8719 \\ & CRISPO & 0.5575 & 0.2941 & 0.4614 & 0.8875 \\ & CRISPO (100 iter) & 0.5513 & 0.2937 & 0.4650 & 0.8888 \\ & ZERA & 0.5493 & 0.2912 & 0.4481 & 0.8862 \\ \rowcolor{midasrow} & MIDAS & \textbf{0.5684} & \textbf{0.3079} & \textbf{0.4749} & \textbf{0.8896} \\ \rowcolor{midasrow} & MIDAS (Multi-LLM) & 0.5574 & 0.2974 & 0.4633 & 0.8857 \\ \cmidrule(lr){1-6} \multirow{7}{*}{\texttt{Format 5}} & ZS & 0.6471 & 0.3671 & 0.6055 & 0.9220 \\ & ICL & 0.7055 & 0.4554 & 0.6721 & 0.9396 \\ & CRISPO & 0.7455 & 0.5409 & 0.7142 & 0.9626 \\ & CRISPO (100 iter) & 0.6717 & 0.4350 & 0.6440 & 0.9272 \\ & ZERA & 0.8009 & 0.5940 & 0.7387 & 0.9363 \\ \rowcolor{midasrow} & MIDAS & \textbf{0.8277} & \textbf{0.6393} & \textbf{0.7716} & 0.9745 \\ \rowcolor{midasrow} & MIDAS (Multi-LLM) & 0.8253 & 0.6276 & 0.7655 & \textbf{0.9746} \\ \bottomrule \end{tabular} \end{table}

Table~\ref{tab:main_results_gpt52} reports evaluation results on the enterprise IT help desk benchmark across all five output configurations. \ac{MIDAS} achieves the strongest overall performance, consistently outperforming all baselines across \texttt{Format~2}--\texttt{Format~5} while remaining competitive with ZERA on \texttt{Format~1}, where it achieves the highest BERTScore F$_1$. These results demonstrate that data-driven policy learning yields consistent gains over both non-optimized baselines and critique-driven optimization frameworks.

\paragraph{Gains over non-optimized baselines.}
Compared to Zero-Shot, \ac{MIDAS} improves \ac{ROUGE}-1 by up to 18.1 points
(\texttt{Format~5}: 0.8277 vs.\ 0.6471) and BERTScore F$_1$ by up to 5.3 points
(\texttt{Format~5}: 0.9745 vs.\ 0.9220). \ac{ICL} narrows this gap but remains
consistently below \ac{MIDAS}, confirming that fixed demonstration examples alone are
insufficient to capture domain-specific formatting constraints.

\paragraph{Gains over critique-driven baselines.}
\ac{MIDAS} consistently outperforms both Cri-SPO and ZERA across
\texttt{Format~2}--\texttt{Format~5}, with the largest gains observed on
\texttt{Format~5}. Compared to \ac{CriSPO}, \ac{MIDAS} improves \ac{ROUGE}-1 from 74.6
to 82.8, \ac{ROUGE}-2 from 54.1 to 63.9, and \ac{ROUGE}-L from 71.4 to 77.2 on this
format. Notably, on \texttt{Format~3} --- the multilingual free-text configuration ---
\ac{CriSPO} underperforms \ac{ICL} (\ac{ROUGE}-1: 0.3749 vs.\ 0.4418), suggesting that
generic critique-driven optimization can regress without data-grounded constraints. In
contrast, \ac{MIDAS} achieves a \ac{ROUGE}-1 score of 0.4862 on the same format, which we
attribute to the policy block explicitly encoding language and structural conventions
extracted from reference summaries.

\paragraph{\texttt{Format~2} is the hardest task.}
Subject line generation yields the lowest absolute scores across all
methods, reflecting the difficulty of compressing a full ticket into a single concise line
with high lexical fidelity. Despite this, \ac{MIDAS} still outperforms all baselines,
including critique-driven optimization frameworks such as \ac{CriSPO} and ZERA,
suggesting that iterative prompt refinement grounded in reference patterns is particularly
beneficial for constrained, short-form generation.

\paragraph{ Architectural Advantage. }

Extending \ac{CriSPO} to 100 iterations (based on best results from
\cite{He2024CriSPOMC}) yields diminishing returns and in some configurations produces
lower scores than its 30-iteration counterpart (e.g., \texttt{Format~5}:
\ac{ROUGE}-1 0.6717 vs.\ 0.7455), suggesting that additional optimization steps alone are
insufficient without data-grounded constraints to guide refinement. Similarly, although
ZERA introduces more sophisticated critique dimensions, its task-agnostic refinement
strategy still underperforms \ac{MIDAS} across most formats. These findings highlight the importance of \ac{MIDAS}'s policy block and unified
\ac{CoT} critic, which provide reference-informed structural grounding
rather than relying solely on longer optimization or generic critique refinement.

\paragraph{Effect of Optimization Iterations. }

\begin{figure}[htbp]
    \centering
    \includegraphics[width=\textwidth]{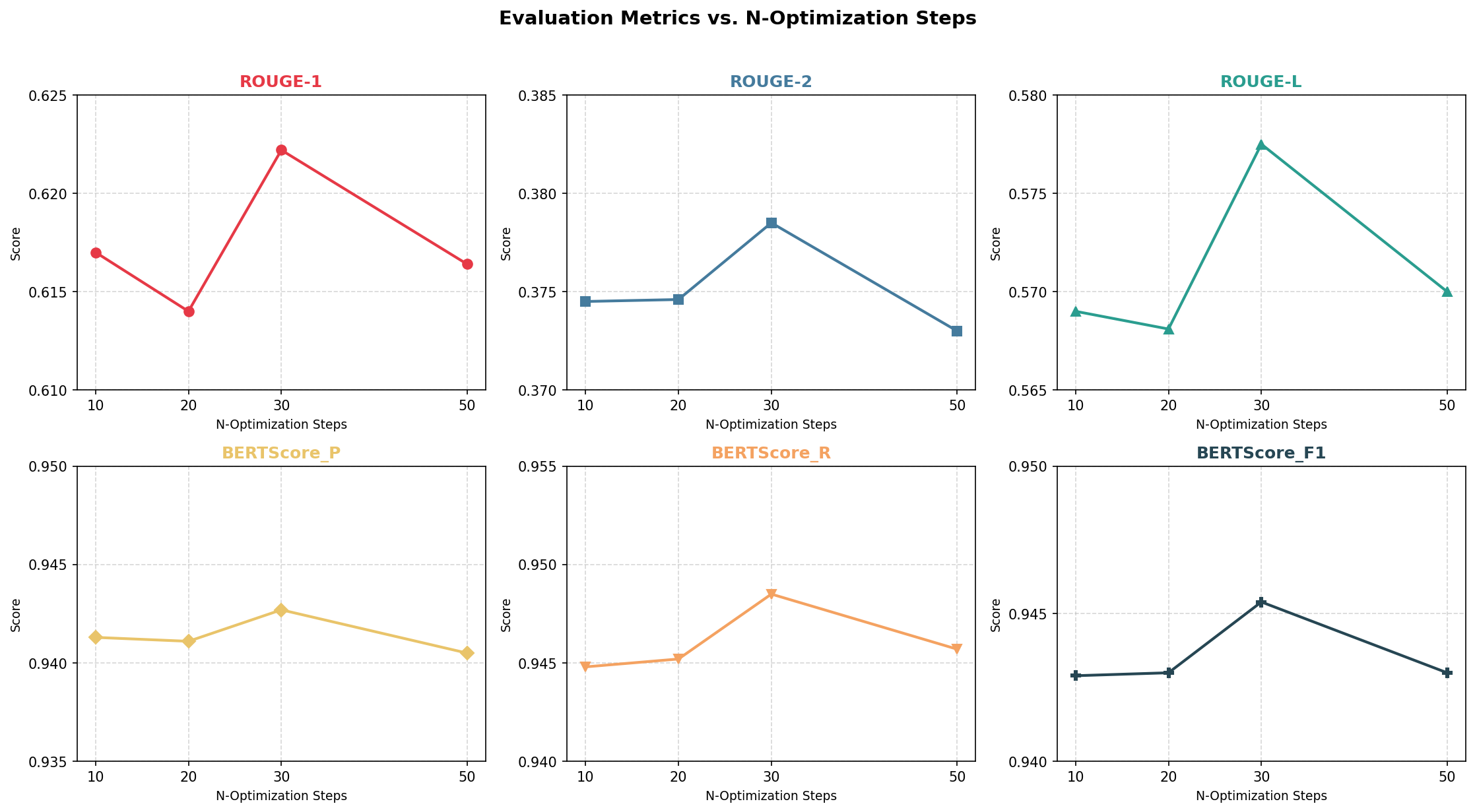}
    \caption{Effect of optimization iterations on summarization quality. 
    \ac{ROUGE}-1/2/L and BERTScore P/R/F$_1$ across $N \in \{10, 20, 30, 50\}$ 
    iterations on a 30-sample test set. $N{=}30$ achieves the best
    performance across all metrics.}
    \label{fig:opt_plot}
\end{figure}

Figure~\ref{fig:opt_plot} shows evaluation metrics across $N \in \{10, 20, 30, 50\}$
optimization steps. Performance improves from $N{=}10$ to $N{=}30$ across all six metrics
(\ac{ROUGE}-1/2/L and BERTScore P/R/F$_1$), then declines slightly at $N{=}50$. This
pattern is consistent across metrics, indicating that $N{=}30$ represents the optimal
performance-efficiency trade-off. All reported results use $N{=}30$.

\paragraph{Generalization Across Output Types. }

\ac{MIDAS} achieves top overall performance across all five structurally distinct output configurations,
ranging from categorical metadata (\texttt{Format~1}) to free-text multilingual summaries
(\texttt{Format~3}) and structured tag outputs (\texttt{Format~5}). The consistent advantage
across formats of varying complexity and output structure demonstrates that \ac{MIDAS}'s
data-driven pattern learning generalizes beyond any single output type, adapting its
optimization trajectory to the conventions of each target format without manual
reconfiguration.

\paragraph{Large-Scale Validation. }
Beyond the 30-sample test sets used during optimization, we validate the
best-performing prompts from each configuration on the full 24,355-ticket
holdout corpus from the enterprise IT help desk dataset. Results on this
large-scale set are consistent with optimization-time findings, confirming that MIDAS
generalizes beyond the small optimization pool to large-scale unseen data.

\paragraph{Multi-LLM Generalization. } To evaluate whether \ac{MIDAS} depends on a single backbone model, we construct a heterogeneous multi-LLM configuration in which different agents are instantiated using different foundation models. Specifically, Claude Sonnet 4.6 is used as the Scorer LLM, Claude Opus 4.6 is used for critique generation and data pattern learning, and GPT-5.2 is used for summary generation and prompt optimization. All runs use the same 30-iteration setting identified in Figure~\ref{fig:opt_plot}. Table~\ref{tab:main_results_gpt52} compares the standard single-backbone GPT-5.2 configuration against a heterogeneous multi-LLM variant of \ac{MIDAS}. Results show that \ac{MIDAS} maintains strong performance under heterogeneous model assignments across all output configurations, confirming that the framework is model-agnostic rather than tied to a specific LLM family. Although the single-backbone GPT-5.2 configuration achieves the strongest overall performance, the multi-LLM configuration remains highly competitive and achieves slightly higher scores on several metrics, suggesting that specialization across different LLM agents can benefit iterative prompt optimization.

\paragraph{Cross-Domain Generalization.} 
\begin{table}[htbp] 
\centering 
\caption{Evaluation results on the ECTSum finance-domain benchmark using \textbf{GPT-5.2} and \textbf{30 optimization iterations}. Best result per metric is shown in \textbf{bold}.} \label{tab:finance_results} \begin{tabular}{@{} l cccc @{}} \toprule \textbf{Method} & \textbf{R-1} & \textbf{R-2} & \textbf{R-L} & \textbf{BS-F1} \\ \midrule CRISPO & 0.3521 & 0.1859 & 0.3078 & 0.8278 \\ ZERA & 0.3630 & 0.1816 & 0.2716 & 0.8217 \\ MIDAS & \textbf{0.3953} & \textbf{0.2414} & \textbf{0.3454} & \textbf{0.8508} \\ \bottomrule \end{tabular} 
\end{table} 

Table~\ref{tab:finance_results} reports results on the ECTSum finance-domain benchmark. MIDAS
achieves the best performance across all evaluation metrics, outperforming
both CriSPO and ZERA despite the substantial domain shift from enterprise IT
help desk summarization to finance-oriented summarization. These results
suggest that MIDAS learns transferable structural constraints rather than
overfitting to a single domain.

\section{Conclusion} \label{sec:conclusion} We presented \ac{MIDAS}, a multi-\ac{LLM} framework for enterprise summarization that extends \ac{CriSPO} with data-driven policy learning and a unified \ac{CoT} critic. By extracting domain-specific formatting constraints from reference summaries and grounding iterative prompt refinement in these patterns, \ac{MIDAS} adapts to diverse summarization requirements without manual prompt engineering. Across five output configurations on a 24,635-ticket enterprise IT help desk corpus, MIDAS achieves the strongest overall performance, consistently outperforming zero-shot, ICL, CriSPO, and ZERA across Format 2--Format 5 while remaining competitive with ZERA on Format 1. Compared to CriSPO, MIDAS improves ROUGE-1 by up to 11.0\%, ROUGE-2 by up to 18.2\%, and ROUGE-L by up to 8.0\%, with gains that generalize across large-scale validation, heterogeneous multi-LLM configurations, and finance-domain benchmarks. 
\paragraph{\textbf{Limitations and Future Work:}} Further evaluation on smaller open-weight models and broader enterprise domains would strengthen deployment robustness claims. Future work may also explore dynamic policy adaptation under evolving organizational conventions.

\begin{credits}

\subsubsection{\discintname}
The authors declare no competing interests relevant to this work.
\end{credits}

\bibliographystyle{splncs04}
\bibliography{refs}

\appendix
\section{Prompt Templates}
\label{app:prompts}

Appendix A presents representative prompt templates that preserve the core structure and optimization behavior used in MIDAS while omitting implementation-specific verbosity for readability.

\lstdefinestyle{promptstyle}{
  basicstyle=\ttfamily\footnotesize,
  columns=fullflexible,
  breaklines=true,
  breakatwhitespace=true,
  frame=none,
  showstringspaces=false,
  keepspaces=true,
  upquote=true,
  aboveskip=2pt,
  belowskip=2pt
}

\subsection{Initial Prompt}
\label{app:init_prompt}

\begin{lstlisting}[style=promptstyle]
INITIAL_PROMPT = "Provide summary following the format of reference summary."
\end{lstlisting}

\subsection{Data Pattern LLM Template (Policy Block Induction)}
\label{app:data_pattern_prompt}

\begin{tcolorbox}[
  enhanced,
  breakable,
  colback=white,
  colframe=black,
  boxrule=0.4pt,
  arc=1pt,
  left=2pt,
  right=2pt,
  top=2pt,
  bottom=2pt,
  boxsep=1pt,
  before skip=2pt,
  after skip=2pt
]
\begin{lstlisting}[style=promptstyle]
_POLICY_BLOCK_INSTRUCTION = """
Analyze input-output pairs to discover transformation patterns.
Generate actionable rules describing HOW inputs become outputs.

CONSTRAINTS:
- Rules must describe observable output patterns.
- Do NOT refer to references during inference.
- No fallback/default outputs.
- Use affirmative, measurable instructions.
- Include ALL distinct feedback rationales when provided.

ANALYZE:
- Formatting patterns
- Structural constraints
- Length patterns
- Domain terminology

FORMAT:
"[When input characteristic X] -> [transformation Y] + [Example]"

OUTPUT:
Only numbered rules. No preamble.
""".strip()
\end{lstlisting}
\end{tcolorbox}

\subsection{Implicit Chain-of-Thought Critic and Prompt LLM Template}
\label{app:critic_prompt_template}

\begin{tcolorbox}[
  enhanced,
  breakable,
  colback=white,
  colframe=black,
  boxrule=0.4pt,
  arc=1pt,
  left=2pt,
  right=2pt,
  top=2pt,
  bottom=2pt,
  boxsep=1pt,
  before skip=2pt,
  after skip=2pt
]
\begin{lstlisting}[style=promptstyle]
_P_AUTOMATIC = """
You are an expert prompt engineer.
Analyze the current prompt and propose an improved standalone task prompt.

GOAL:
Ensure outputs satisfy:
- Structural inconsistencies
- Content drift
- Structural format
- Style consistency

REQUIREMENTS:
1) Provide critique inside <Critique></Critique>
2) Provide FULL revised prompt inside <Suggestion></Suggestion>
3) Output only these two sections
4) New prompt must be self-contained
5) Use input-based rules (no reference comparisons)
6) Avoid fallback/default outputs
7) Respect typical target length

{policy_block}

INPUTS:
- Current prompt + score
- Generated examples
- Optimization history

ADDRESS:
- Reference-dependent wording
- Prefix hallucinations
- Content drift
- Length mismatch
""".strip()
\end{lstlisting}
\end{tcolorbox}

\subsection{Scorer LLM Template}
\label{app:scorer_prompt}

\begin{tcolorbox}[
  enhanced,
  breakable,
  colback=white,
  colframe=black,
  boxrule=0.4pt,
  arc=1pt,
  left=2pt,
  right=2pt,
  top=2pt,
  bottom=2pt,
  boxsep=1pt,
  before skip=2pt,
  after skip=2pt
]
\begin{lstlisting}[style=promptstyle]
LLM_EVALUATION_PROMPT_TEMPLATE = """
Evaluate generated output against target output.

Score:
1) Core Meaning: Semantic correctness and preservation of key information
2) Unsupported Additions: Penalize hallucinated or unsupported content
3) Format & Style Fidelity: Adherence to structural and stylistic constraints

Return:
Core Meaning: [0-100]
Unsupported Additions: [0-100]
Format & Style Fidelity: [0-100]
Explanation: [...]
""".strip()
\end{lstlisting}
\end{tcolorbox}

\end{document}